\pdfoutput=1

\documentclass[11pt]{article}

\usepackage[final]{ACL2023}

\usepackage{fdsymbol}
\usepackage{graphicx}

\usepackage[T1]{fontenc}

\usepackage[utf8]{inputenc}

\usepackage{microtype}

\usepackage{inconsolata}

\usepackage{times}
\usepackage{danudefs}

\usepackage{xcolor}
\definecolor{OliveGreen}{RGB}{0,102,0}

\usepackage{booktabs}
\usepackage{xspace}
\usepackage{url}

\usepackage[english]{babel}
\usepackage{hyperref}
\addto\captionsenglish{%
}
\addto\extrasenglish{%
}

\title{Solving Cosine Similarity Underestimation between \\High Frequency Words by $\ell_2$ Norm Discounting}

\author{Saeth Wannasuphoprasit$^\spadesuit$ \And Yi Zhou$^\diamondsuit$ \\
  University of Liverpool$^\spadesuit$, Cardiff University$^\diamondsuit$, Amazon$^\clubsuit$\\
 {\tt \{s.wannasuphoprasit,danushka\}@liverpool.ac.uk} \\ 
 {\tt zhouy131@cardiff.ac.uk} \And
 Danushka Bollegala$^{\clubsuit,\spadesuit}$}

\date{}

\begin{document}
\maketitle

\begin{abstract}
Cosine similarity between two words, computed using their contextualised token embeddings obtained from masked language models (MLMs) such as BERT has shown to underestimate the actual similarity between those words~\cite{zhou-etal-2022-problems}.
This similarity underestimation problem is particularly severe for highly frequent words.
Although this problem has been noted in prior work, no solution has been proposed thus far.
We observe that the $\ell_2$ norm of contextualised embeddings of a word correlates with its log-frequency in the pretraining corpus.
Consequently, the larger $\ell_2$ norms associated with the highly frequent words reduce the cosine similarity values measured between them, thus underestimating the similarity scores.
To solve this issue, we propose a method to \emph{discount} the $\ell_2$ norm of a contextualised word embedding by the frequency of that word in a corpus when measuring the cosine similarities between words.
We show that the so called \emph{stop} words behave differently from the rest of the words, which require special consideration during their discounting process.
Experimental results on a contextualised word similarity dataset show that our proposed discounting method accurately solves the similarity underestimation problem.
\end{abstract}

\section{Introduction}
\label{sec:intro}

Cosine similarity is arguably the most popular word similarity measure used in numerous natural language processing (NLP) tasks, such as question answering (QA), information retrieval (IR) and machine translation (MT)~\cite{echizen2019word, oniani2020qualitative, Kim:2021, hanifi2022problem}.
First, a word is represented by a vector (aka \emph{embedding}) and then the similarity between two words is computed as the cosine of the angle between the corresponding vectors~\cite{rahutomo2012semantic}.
Despite the good performance of cosine similarity as a similarity measure in various downstream tasks, \citet{zhou-etal-2022-problems} showed that it systematically underestimates the true similarity between highly frequent words, when computed using contextualised word embeddings obtained from MLMs such as BERT~\cite{BERT}.

Compared to the problem of estimating similarity between highly frequent words, the opposite problem of estimating the similarity between (or involving) rare (low frequency) words has received greater attention, especially in the scope of static word embeddings~\cite{levy2014neural,hellrich-hahn-2016-bad,mimno-thompson-2017-strange,wendlandt-etal-2018-factors}. 
If a word is rare in a corpus, we might not have a sufficiently large number of contexts containing that word to learn an accurate embedding for it.
This often leads to unreliable similarity estimations between words and has undesirable implications in downstream tasks such as the detection of analogies and social biases~\cite{ethayarajh-etal-2019-towards, ethayarajh2019understanding}.

On the other hand, \citet{zhou-etal-2022-problems} studied the impact of frequency on contextualised word embeddings and showed that the cosine similarity between highly frequent words are systematically underestimated.
Unlike in the previously discussed low frequency word scenario, we do have adequate contexts to learn an accurate semantic representation for highly frequent words.
Therefore, it might appear surprising at first that cosine similarity cannot be correctly estimated even for the highly frequent words.
\citet{zhou2021frequency} show that the diversity (measured by the volume of the bounding hypersphere)  of the contextualised embeddings of a target word, computed from multiple contexts containing the word, increases with the frequency of that word.
They provide an explanation that holds true only for 2-dimensional embeddings, which relates diversity to the underestimation of cosine similarity.
Unfortunately, this explanation does not extend to the high dimensional embeddings used in practice by the NLP community (e.g. BERT token embeddings are typically more than 768 dimensional).
More importantly, to the best of our knowledge, no solution has been proposed in the literature to address the cosine similarity underestimation problem associated with the highly frequent words.

\begin{figure}[t]
    \centering
    \includegraphics[width=\linewidth]{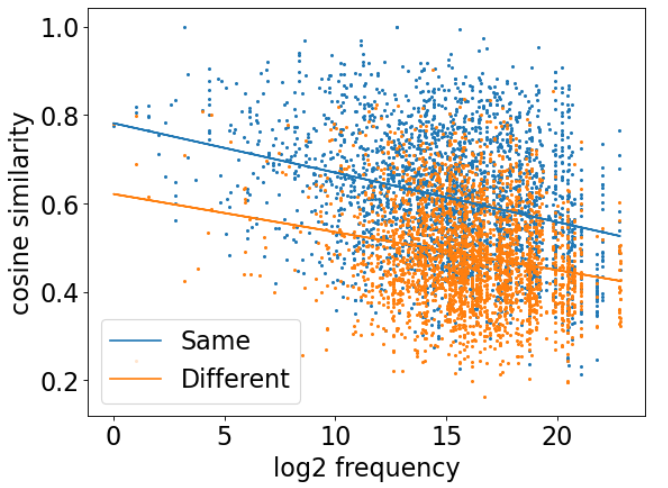}
    \caption{Cosine similarity between two instances of the same word $w$ in two contexts in the WiC train dataset. When the log-frequency of $w$ in the corpus increases, cosine similarities computed for both contexts that express the same meaning of $w$ as well as its different meanings decreases.
    }
    \label{fig:scatter-original}
\end{figure}

In prior work, the $\ell_2$ norm of a static word embedding has been shown to linearly correlate with the log-frequency of that word~\cite{arora-etal-2016-latent,Bollegala:AAAI:2018}.
On the other hand, we empirically study the $\ell_2$ norm of the contextualised embedding of a word $w$ averaged over all of its contexts, and find that it too approximately linearly correlates with the log-frequency of $w$ in the corpus used to pretrain the MLM.
Recall that the cosine similarity is defined as the inner-product between two embeddings, divided by the $\ell_2$ norm of those embeddings.
Therefore, we suspect that the underestimation of cosine similarity between highly frequent words is due to the larger $\ell_2$ norms associated with those words.

To correct for this bias associated with the $\ell_2$ norms of highly frequent words, we propose a linearly parameterised discounting scheme in the log-frequency space.
Specifically, we use Monte-Carlo Bayesian Optimisation~\cite{botorch} to find the optimal discounting parameters.
Our proposed discounting method is shown to accurately correct the underestimation of cosine similarities between highly frequent words on the Word-in-Context (WiC)~\cite{Pilehvar:2019} dataset where human similarity ratings are available for the same word in two different contexts.
Source code for reproducing the experiments reported in this is paper is publicly available.\footnote{\url{https://github.com/LivNLP/cosine-discounting}}

\section{Underestimation of Cosine Similarity}
\label{sec:problem}

Let us denote the $d$-dimensional contextualised word embedding produced by an MLM $f$ for a target word $w$ appearing in a context $c$ by $\vec{f}(w,c) (\in \R^d)$.
Moreover, let the set of contexts where $w$ occurs in a given corpus be $\cS(w)$.
We refer to $\{\vec{f}(w,c) | w \in \cS(w)\}$ as the set of \emph{sibling embeddings} of $w$.
To study the relationship between the cosine similarity scores and the frequency of words, we use the 768-dimensional \texttt{bert-base-uncased}\footnote{\url{https://huggingface.co/bert-base-uncased}} as the contextualised embedding model.
We use the token embedding of $w$ from the final hidden layer of BERT as $\vec{f}(w,c)$.
We approximate the word frequencies in BERT pretraining corpus using the BookCorpus~\cite{Zhu_2015_ICCV}.
Let $\psi_w$ be the frequency of $w$ in this corpus.

We use the WiC dataset, which contains 5428 pairs of words appearing in various contexts with annotated human similarity judgements.
WiC dataset is split into official training and development sets, while a separate hidden test set is used by the leaderboard for ranking Word Sense Disambiguation systems.\footnote{\url{https://pilehvar.github.io/wic/}}
WiC dataset contains pairs of contexts labelled as having the \textbf{same meaning} (e.g. ``to \emph{drive} sheep out of a field'' vs. ``to \emph{drive} the cows into the barn'') and \textbf{different meaning} (e.g. ``the \emph{play} lasted two hours'' vs. ``they made a futile \emph{play} for power'').

\begin{figure}[t]
    \centering
    \includegraphics[width=\linewidth]{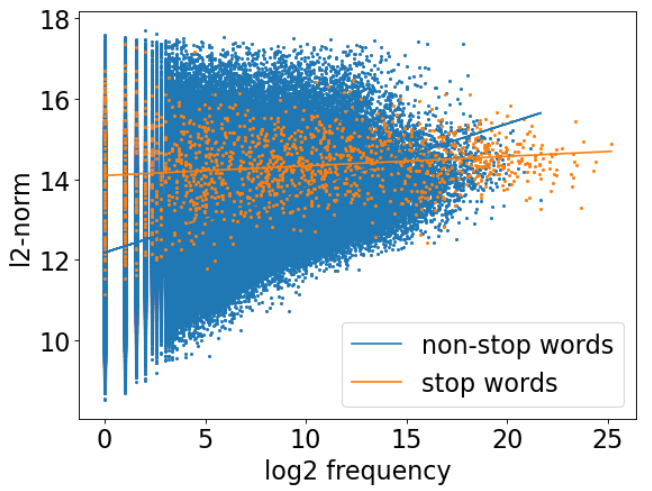}
    \caption{$\ell_2$ norm of the averaged contextualised word embedding of a word against its log-frequency in the pretrain corpus. Stop words and non-stop words are shown respectively in \textcolor{orange}{orange} and \textcolor{blue}{blue} dots. Lines of best fits for each category are superimposed.}
    \label{fig:norm-freq}
\end{figure}

\begin{figure}[h]
\centering
\includegraphics[width=7.5cm]{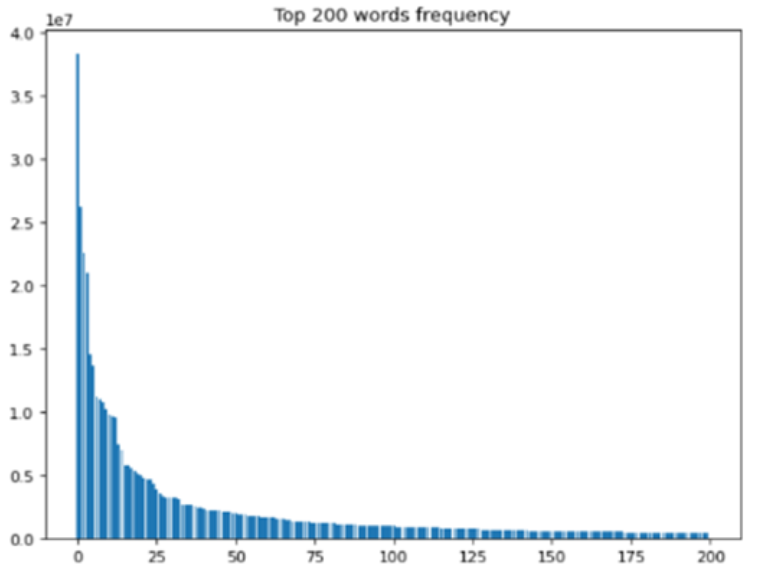}
\caption{Histogram of word frequencies in the BERT pretrain corpus. We see a Zipfian (power-law) distribution, which turns out to be approximately liner in the log-frequency space.}
\label{fig:hist}
\end{figure}

We compute the cosine similarity between the two contextualised embeddings of a target word in two of its contexts to predict a similarity score.
\autoref{fig:scatter-original} shows the predicted similarity scores for both contexts in which a target word has been used in the same or different meanings for all words in the WiC dataset against $\log(\psi_w)$.
As seen from \autoref{fig:hist}, $\psi_w$ has a power-law distribution. 
Therefore, we plot its log instead of raw frequency counts in \autoref{fig:scatter-original}.

From \autoref{fig:scatter-original}, we see that for both same as well as different meaning contexts, the predicted cosine similarities drop with the word frequencies.
Moreover, the gradient of the drop for same meaning pairs (Pearson's $r=-0.3001$) is larger than that for the different meaning pairs ($r=-0.2125$), indicating that the underestimation of cosine similarity is more sever for the similar contexts of highly frequent words.

\section{$\ell_2$ norm Discounting}
\label{sec:discount}

To understand the possible reasons behind the cosine similarity underestimation for highly frequent words discussed in \autoref{sec:problem}, for each word $w$ we compute its mean sibling embedding, $\hat{\vec{w}}$, given by \eqref{eq:mean}.
\begin{align}
	\label{eq:mean}
	\hat{\vec{w}} = \frac{1}{|\cS(w)|} \sum_{c \in \cS(w)} \vec{f}(w,c)
\end{align}
We plot $\norm{\hat{\vec{w}}}$ against $\log(\psi(w))$ in \autoref{fig:norm-freq} separately for a predefined set of stop words and all other words (i.e. non-stop words).
For this purpose, we use the default 1466 stop words from NLTK and randomly selected 997,425 non-stop words from the BookCorpus.
Pearson $r$ values of stop words and non-stop words are respectively 0.1697 and 0.3754, while the lines of best fits for each class of words are superimposed.
From \autoref{fig:norm-freq}, we see that overall, $\norm{\hat{\vec{w}}}$ increases with $\log(\psi_w)$ for both stop and non-stop words, while the linear correlation is stronger in the latter class.
Considering that stop words cover function words such as determiners and conjunctions that co-occur with a large number of words in diverse contexts, we believe that the $\ell_2$ norm of stop words mostly remains independent of their frequency.
Recall that the cosine similarity between two words is defined as the fraction of the inner-product of the corresponding embeddings, divided by the product of the $\ell_2$ norms of the embeddings.
Therefore, even if the inner-product between two words remain relatively stable, it will be divided by increasingly larger $\ell_2$ norms in the case of highly frequent words.
Moreover, this bias is further amplified when both words are high frequent due to the \emph{product} of $\ell_2$ norms in the denominator.

To address this problem, we propose to discount the $\ell_2$ norm of a word $w$ by a discounting term, $\alpha(\psi_w)$, and propose a discounted version of the cosine similarity given by \eqref{eq:discount-cosine}.
\begin{align}
	\label{eq:discount-cosine}
	\cos_{\alpha}(\vec{x}, \vec{y}) = \frac{\vec{x}\T\vec{y}}{\norm{\vec{x}}\alpha(\psi_x) \norm{\vec{y}}\alpha(\psi_y)}
\end{align}

Following \autoref{fig:norm-freq}, we linearly parameterise $\alpha(\psi_w)$ separately for stop vs. non-stop words as in \eqref{eq:alpha}.
\par\nobreak
{\small
\begin{align}
\alpha(\psi_w) = 
		\begin{cases}
		1 + m_s (b_s - \log(\psi_w)) & \text{w is a stop word} \\
		1 + m_n (b_n - \log(\psi_w)) & \text{w is a non-stop word}
		\end{cases}
	\label{eq:alpha}
\end{align}
}%

The scalar parameters $m_s, m_n, b_s$ and $b_n$ are estimated as follows.
First, we randomly initialise all parameters uniformly in $[0,1]$ and use \eqref{eq:discount-cosine} to predict cosine similarity between two contexts in which a target word $w$ occurs in the WiC train instances.
We then make a binary similarity judgement (i.e. \textbf{same} or \textbf{different} meaning) for the pair of contexts in an instance depending on whether the predicted cosine similarity is greater than a threshold $\theta$.
Next, we compute the overall binary classification accuracy for the similarity predictions made on the entire WiC training dataset, and use Bayesian Optimisation to find the optimal values: $\theta = 0.545$, $m_s = 0.00422$, $b_s = 0.643$, $m_n = 0.00427$ and $b_n = 4.821$.
Specifically we used the Adaptive Experimentation Platform\footnote{\url{https://ax.dev/}} for learning those optimal values.
We found this is more efficient than conducting a linear search over the parameter space.
We repeat the estimation five times and use the averaged parameter values in the remainder of the experiments.
Note that $m_n > m_s$ above, which indicates that non-stop words must be discounted slightly more heavily than the stop words.
This makes sense since the impact of word frequency of non-stop words on their $\ell_2$-norm is stronger than that for the stop words as indicated by the slopes of the lines of best fit in \autoref{fig:norm-freq}.

\section{Results}
\label{sec:res}

\begin{figure}[t!]
    \centering
    \includegraphics[width=\linewidth]{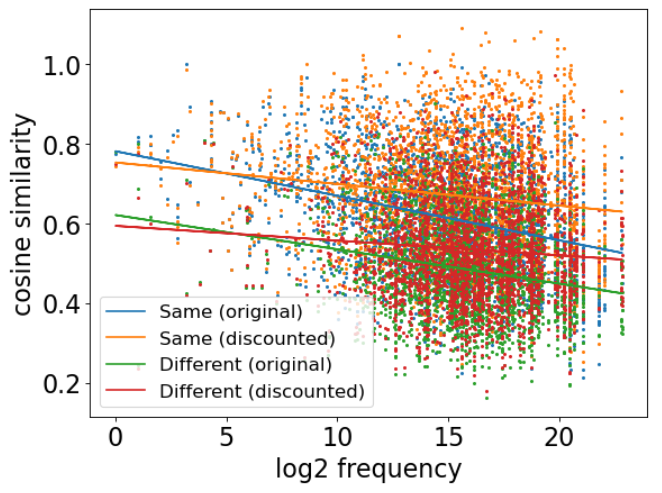}
    \caption{Cosine similarity between two instances of the same word $w$ in two contexts in the WiC train dataset, computed using the original (non-discounted) cosine similarity (shown in \textcolor{blue}{blue} and \textcolor{OliveGreen}{green} respectively for the same and different meaning pairs) and using the proposed $\ell_2$ norm discounted (\eqref{eq:discount-cosine}) (shown in \textcolor{orange}{orange} and \textcolor{red}{red} respectively for the same and different meaning pairs). We see that the gradients of the drops have \emph{decreased} for both same and different meaning pairs \emph{after} applying the discounting.
    }
    \label{fig:scatter-discounted}
\end{figure}

To evaluate the effect of the proposed $\ell_2$ norm discounting when computing cosine similarity, we repeat the analysis presented in \autoref{fig:scatter-original} using \eqref{eq:discount-cosine} to predict the similarity between contextualised word embeddings.
Comparing the lines of best fit for the original (\textcolor{blue}{blue}, $r = -0.3006$) vs. discounted (\textcolor{orange}{orange}, $r=-0.1366$) for the same meaning contexts, we see that the gradient of the drop has decreased by $51.65\%$.
Likewise, comparing the lines of best fit for the original (\textcolor{OliveGreen}{green}, $r=-0.2125$) vs. discounted (\textcolor{red}{red}, $r=-0.0843$) for the different meaning contexts, we see the gradient of the drop has decreased by $57.04\%$.
This result clearly shows that the proposed $\ell_2$ norm discounting method is able to reduce the underestimation of cosine similarities for the highly frequent words.

\begin{figure}[t]
    \centering
    \includegraphics[width=\linewidth]{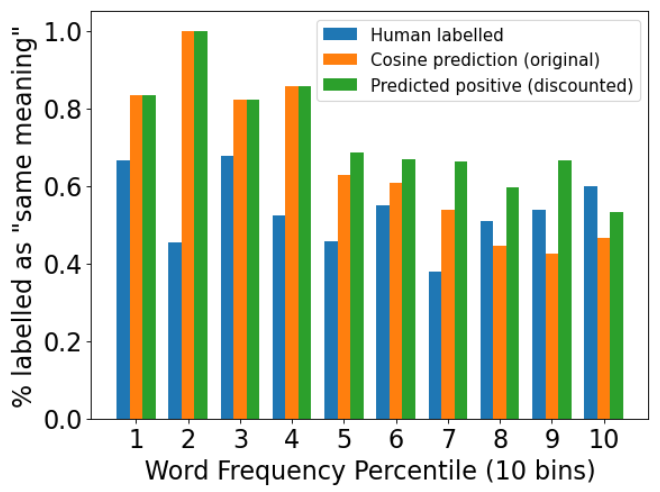}
    \caption{Percentage of examples labelled as having the ``same meaning''. In high frequency words, we see that the cosine similarity-based predictions (\textcolor{orange}{orange}/middle) are systematically \textbf{underestimate} the human similarity judgements (\textcolor{blue}{blue}/left). However, after the proposed discounting method has been applied (\textcolor{OliveGreen}{green}/right) the underestimation has reduced. 
    }
    \label{fig:prediction-discounted}
\end{figure}

Given that the discounting parameters in \eqref{eq:alpha} are learned from the WiC train data, it remains an open question as to how well the proposed discounting method generalises when predicting similarity between contextualised embeddings of unseen words.
To evaluate this generalisability of the proposed method, we use \eqref{eq:alpha} with its learned parameters from WiC train data, to predict the similarity between contextualised word embeddings in WiC dev data.\footnote{Note that the test set of WiC is publicly unavailable due to being used in a leaderboard.}
Specifically, we predict binary (same vs. different meaning) similarity labels according to the similarity threshold $\theta$ learnt in \autoref{sec:discount} and compare against the  human judgements using binary classification accuracy.

The maximum accuracy on WiC dev split obtained using the original (non-discounted) cosine similarities is $0.6667$, which indicates that the cosine similarity is somewhat predictive of the human binary judgements.
The overall F1 is improved by $2.4\%$ (0.68 with original cosine vs. 0.71 with the proposed discounting method) and recall is improved by $12\%$ (0.75 with original cosine vs. 0.84 with the proposed). On the other hand, the drop in precision is $4.7\%$ (from 0.64 to 0.61). 
Therefore, the proposed method solves the cosine similarity underestimation problem associated with high-frequent words, without significantly affecting the similarity scores for low-frequent ones

\autoref{fig:prediction-discounted} shows the average proportion of instances predicted to be the same meaning as a function of frequency, grouped into ten bins, each with the same number of examples.
From \autoref{fig:prediction-discounted}, we see that in high frequency bins (i.e. bins 8, 9 and 10), the percentage of predicted instances as having the same meaning is consistently lower than that compared to the human judgements.
This shows an underestimation of the true (human judged) similarity between contextualised word embeddings.

On the other hand, when we use the proposed $\ell_2$ norm discounted cosine similarity (defined in \eqref{eq:discount-cosine}), in the highest frequent bin (i.e. 10) we see that the gap between human judgements vs. predicted similarities has reduced.
Moreover, in the low frequency bins (i.e. 1--4), we see that the proposed discounting method does not affect the predictions made using cosine similarities.
We see an overestimation of the cosine similarities in the low frequency bins as reported by \citet{zhou2021frequency}.
As discussed already in \autoref{sec:intro}, the word embeddings learnt for low frequency words tend to be unreliable due to data sparseness.
Therefore, we believe it is important to focus on the problem of learning accurate word embeddings rather than to adjust cosine similarities between low-frequency words in a post-processing step.

We see that in bins 5, 6 and 7 the similarity scores are slightly increased by the proposed discounting method, which is a drawback that needs to be addressed in future work.
More importantly however, the overall percentage recall across all bins for retrieving same meaning instances improves significantly from 74.7\% to 83.7\% compared to using respectively the original cosine similarity vs. the discounted cosine similarity.
Overall, this result confirms the validity of the proposed discounting method for addressing the underestimation of cosine similarity involving highly frequent words.

\section{Conclusion}

We proposed a method to solve the cosine similarity underestimation problem in highly frequent words.
Specifically, we observed that the $\ell_2$ norm of a contextualised word embedding increases with its frequency in the pretrain corpus and proposed a discounting scheme.
Experimental results on WiC dataset confirmed the validity of the proposed method.

\section{Limitations}

We proposed a solution to the cosine similarity underestimation problem associated with contextualised word embeddings of highly frequent words.
Our evaluations used only a single contextualised embedding model (i.e. BERT) with a single dimensionality (i.e. 768).
Therefore, we believe that our proposed method must be evaluated with other (more recent) MLMs to test for its generalisability.
Moreover, our evaluations were conducted only on the English language, which is known to be morphologically limited.
Although in our preliminary experiments we considered discounting schemes based on the part-of-speech of words (instead of considering stop words vs. non-stop words), we did not find any significant improvements despite the extra complexity.
However, these outcomes might be different for more morphologically richer languages.
In order to evaluate similarity predictions in other languages, we must also have datasets similar to WiC annotated in those languages, which are difficult to construct.
Although having stated that using a single MLM and single language as limitations of this work, we would like to point out that these are the same conditions under which \citet{zhou-etal-2022-problems} studied the cosine similarity underestimation problem.

We used only a single dataset (i.e. WiC) in our experiments in this short paper due to space constraints.
Other contextual similarity datasets (e.g. Stanford Contextualised Word Similarity (SCWS)~\cite{Huang:ACL:2012}) could be easily used to further validate the proposed discounting method in an extended version.

\section{Ethical Considerations}


In this paper, we do not annotate novel datasets nor release any fine-tuned MLMs. 
Therefore, we do not see any direct ethical issues arising from our work.
However, we are proposing a method to address the underestimation of cosine similarity scores computed using contextualised word embeddings obtained from (possibly socially biased) pretrained MLMs.
We would therefore discuss the ethical implication of this aspect of our work in this section.

Cosine similarity has been used in various social bias evaluation measures such as the WEAT~\cite{WEAT}, SemBias~\cite{Zhao:2018aa}, WAT~\cite{du-etal-2019-exploring}, etc.
These methods measure the cosine similarity between a gender and a set of pleasant or unpleasant set of attributes to compute a social bias evaluation score.
Although originally these methods were developed for evaluating the social biases in static word embeddings, they have been later extended to contextualised word embeddings~\cite{Kaneko:2022,kaneko-etal-2022-gender} and sentence embeddings~\cite{may-etal-2019-measuring}, where cosine similarity still remains the main underlying metric.
However, \citet{ethayarajh-etal-2019-understanding} showed that inner-products to be superior over cosine similarity for social bias evaluation purposes.
It remains unclear as to how the underestimation in cosine similarities discussed in our work would influence the social bias evaluations.
In particular, the effect of the proposed $\ell_2$ norm discounting scheme on social bias evaluation must be carefully studied in the future work.

\section*{Acknowledgements}
Danushka Bollegala holds concurrent appointments as a Professor at University of Liverpool and as an Amazon Scholar. This paper describes work performed at the University of Liverpool and is not associated with Amazon.

\bibliography{myrefs.bib}
\bibliographystyle{acl_natbib}

\end{document}